\pdfoutput=1

\documentclass[11pt]{article}

\usepackage[]{ACL2023}

\usepackage{times}
\usepackage{latexsym}
\usepackage{multirow}
\usepackage{graphicx}
\usepackage{array}
\usepackage{adjustbox}
\usepackage{todonotes}
\usepackage{pgfplots} 
\usepackage{makecell}
\usepackage{pbox}

\usepackage[T1]{fontenc}

\usepackage[utf8]{inputenc}

\usepackage{microtype}

\makeatletter
\def\thickhline{%
  \noalign{\ifnum0=`}\fi\hrule \@height \thickarrayrulewidth \futurelet
   \reserved@a\@xthickhline}
\def\@xthickhline{\ifx\reserved@a\thickhline
               \vskip\doublerulesep
               \vskip-\thickarrayrulewidth
             \fi
      \ifnum0=`{\fi}}
\makeatother

\newlength{\thickarrayrulewidth}
\makeatletter
\newcommand{\printfnsymbol}[1]{%
  \textsuperscript{\@fnsymbol{#1}}%
}
\makeatother

\usepackage{inconsolata}

\title{Adapting Text-based Dialogue State Tracker for Spoken Dialogues}

\author{
    Jaeseok Yoon\textsuperscript{\rm 1}\thanks{\ \ These authors contributed equally.} , 
    Seunghyun Hwang\textsuperscript{\rm 1}\printfnsymbol{1},
    Ran Han\textsuperscript{\rm 2},
    Jeonguk Bang\textsuperscript{\rm 2}, 
    Kee-Eung Kim\textsuperscript{\rm 1,\rm 3} \\
    \textsuperscript{\rm 1}Kim Jaechul Graduate School of AI, KAIST, Seoul, Republic of Korea\\
    \textsuperscript{\rm 2}Electronics Telecommunications Research Institute (ETRI), Daejeon, Republic of Korea\\
    \textsuperscript{\rm 3}School of Computing, KAIST, Daejeon, Republic of Korea\\ 
    \{jake.yoon, steven1971\}@kaist.ac.kr, \{ran.han, jubang0219\}@etri.re.kr, kekim@kaist.ac.kr 
}

\begin{document}
\maketitle
\begin{abstract}

%
Although there have been remarkable advances in dialogue systems through the dialogue systems technology challenge (DSTC), it remains one of 
the key challenges to building a robust task-oriented dialogue system with a speech interface. Most of the progress has been made for text-based dialogue systems
since there are abundant datasets with written corpora while those with spoken dialogues are very scarce. However, as can be seen from voice assistant systems such
as Siri and Alexa, it is of practical importance to transfer the success to spoken dialogues. In this paper, we describe our engineering effort in building a highly
successful model that participated in the speech-aware dialogue systems technology challenge track in DSTC11. Our model consists of three major modules:
(1) automatic speech recognition error correction to bridge the gap between the spoken and the text utterances, (2) text-based dialogue system (D3ST) for estimating the slots 
and values using slot descriptions, and (3) post-processing for recovering the error of the estimated slot value. Our experiments show that it is important
to use an explicit automatic speech recognition error correction module, post-processing, and data augmentation to adapt a text-based dialogue state tracker for spoken dialogue corpora. 

\end{abstract}

\section{Introduction}
Task-oriented dialogue (TOD) systems aim to successfully perform various tasks, such as hotel reservations, train information retrieval, and food ordering, by interacting with users. 
In particular, with the popularity of voice assistant systems such as Siri and Alexa, it is important to be able to take spoken languages as inputs to TOD systems. These TOD systems with the speech interface is referred to as \textit{spoken dialogue systems}.

However, due to the differences in many aspects between text conversation and speech conversation, the existing TOD systems for text conversation can fail to achieve good performance for speech conversation. To this end, several tasks such as ATIS \citep{hemphill1990atis} and competitions (DSTC2\footnote{\url{https://github.com/matthen/dstc}}, DSTC10 knowledge-grounded task-oriented dialogue modeling on spoken conversations track\footnote{\url{https://github.com/alexa/alexa-with-dstc10-track2-dataset}}) were proposed. In the knowledge-grounded task-oriented dialogue modeling on spoken conversations track in DSTC10, various methods were applied to spoken dialogue systems. \citet{khan2015hypotheses} introduced a method to reduce automatic speech recognition (ASR) error by making predictions using multiple hypotheses of several ASR models together. Also, \citet{tian2021tod} proposed a method to reduce the discrepancy between written conversation and spoken conversation by augmenting the dataset based on ontology and pattern or generating noisy addition of spoken conversation through a text-to-speech (TTS)->ASR pipeline. 

\begin{table*}[h!]
    \begin{adjustbox}{width=\textwidth}
    \begin{tabular}[t]{@{}l|l|l}
        \hline
        \textbf{Speaker}&\textbf{Utterance}&\textbf{Dialog state}\\
        \hline
        \hline
        User&\pbox{14cm}{I'd like to find out if there are any \textcolor{red}{4-star} rated \textcolor{red}{guesthouses} in cambridge}&\multirow{10}{*}{\makecell{hotel-stars=4\\
hotel-type=guesthouse\\
hotel-internet=yes\\
restaurant-name=bangkok city\\
restaurant-area=centre\\
}}\\
        System&\pbox{14cm}{There are quite a few. what area do you prefer?}&\\
        User&\pbox{14cm}{I'm open to any area as long as there is \textcolor{red}{free wifi}.}&\\
        System&\pbox{14cm}{Then I recommend the a and b guest house. Would you like me to book you a room?}&\\
        User&\pbox{14cm}{Does that also have free parking available?}&\\
        System&\pbox{14cm}{No, it doesn't. should I recommend you a place with free parking instead?}&\\
        User&\pbox{14cm}{No, but I am looking for a particular restaurant. Its name is called \textcolor{red}{bangkok city}}&\\
        System&\pbox{14cm}{Bangkok city is an expensive Thai restaurant in the \textcolor{red}{centre} of town. They are located at 24 green street city centre. Their postcode is cb23jx. Would you like a reservation?}&\\
        User&\pbox{14cm}{All I needed today was the address, thank you}&\\
        \hline
    \end{tabular}
    \end{adjustbox}
    \caption{ A DST example in MultiWOZ dataset }
    \label{tab:multiwoz_example}
\end{table*}

Since only ASR hypothesis data was provided in knowledge-grounded task-oriented dialogue modeling on spoken conversations track in DSTC10, it was difficult to effectively deal with errors generated in ASR, and most of the studies were focused on data augmentation to make up for the lack of data. On the other hand, in the speech-aware dialogue system technology track in DSTC11, in addition to the ASR hypothesis data, audio, transcripts used to generate the audio and augmented data were provided, allowing a more diverse approach.
For this motivation, we studied how to solve problems caused by the propagation of ASR errors to the model and how to solve errors such as incorrect proper nouns generated by spoken input and participated in the speech-aware dialogue systems technology challenge track in DSTC11.

In this paper, we propose a model, which allows good performance for spoken utterance input as well. Our contributions are summarized as follows: (1) We show that explicit ASR error correction can improve the performance of dialogue systems with spoken corpora as input. (2) We show that post-processing can mitigate errors in words such as proper nouns. (3) We successfully construct a dialogue system that performs well with spoken utterance input.

\section{Related Work}

\subsection{Dialogue State Tracking}

Dialogue state tracking (DST) is one of the components of a task-oriented dialogue system that maps partial dialogues to the dialogue state. It usually extracts the user's goal and intent in the form of a slot-value pair through the user and system dialogue conversation.
As an example in Table~\ref{tab:multiwoz_example}, the DST task is to extract dialogue states such as the value of \textit{guesthouse} in the slot of \textit{hotel-type}  and the value of \textit{bangkok city} in the slot of \textit{restaurant-name} from user's utterance. There are several methods have recently attracted attention in DST tasks. 

\textbf{Dialogue Systems with Description Input} Some works have been proposed which include task descriptions as input, where the descriptions related to the dialog system slot or slot value examples are added as input data \citep{shah2019robust}. \citet{zhao2022description} showed that a language description-driven system shows a better understanding of task specifications, higher performance on state tracking, improved data efficiency, and effective zero-shot transfer to unseen tasks. 

\textbf{Prompting Dialogue System} In the NLP community, it has been shown that large language models such as GPT-3 \citep{brown2020language} and LaMDA \citep{thoppilan2022lamda} can do few-shot learning without fine-tuning. In the dialogue state tracking task, \citet{madotto2020language} applied GPT-2 by priming the model with dialogue state value pair examples. Their model works well on dialogue state tracking with few shot examples without fine-tuning.
\begin{figure*}
    \centering
    \includegraphics[width=\linewidth]{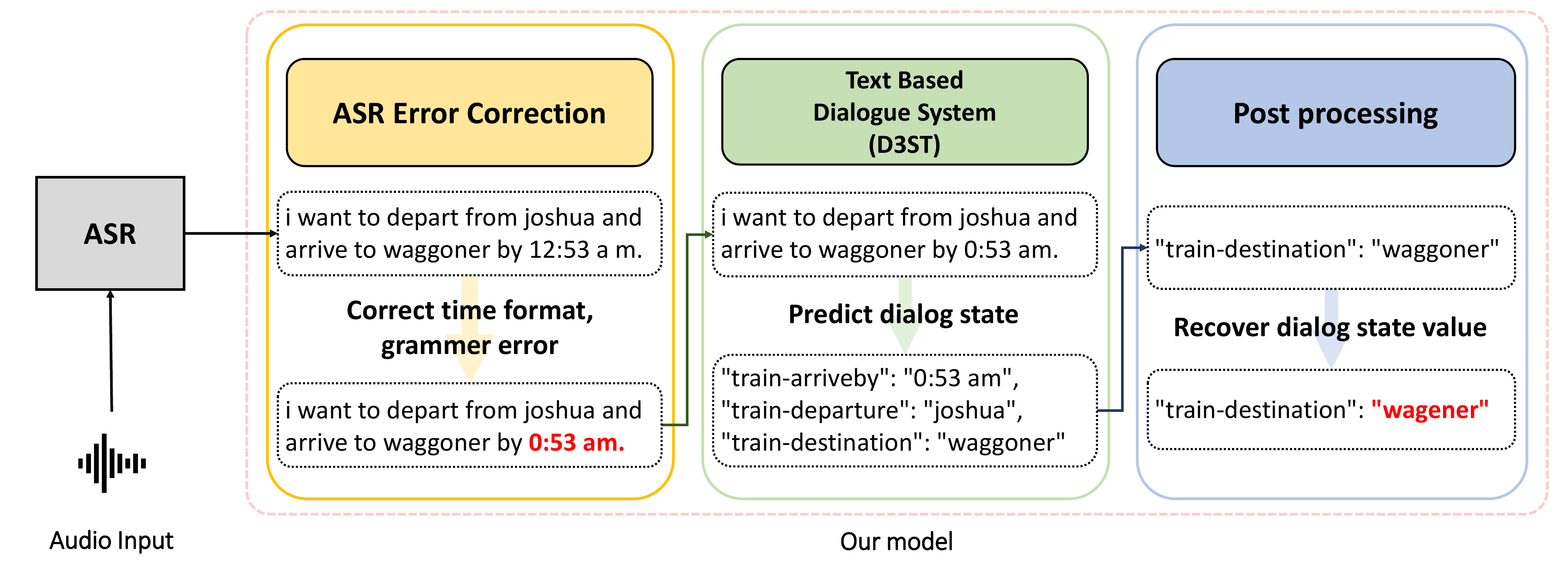}
    \caption{Our model structure}
    \label{fig:model_structure}
\end{figure*}


\subsection{Spoken Dialogue System}

In spoken dialogue systems, the discrepancy between text and speech conversations makes it difficult for text-based dialogue systems to perform well. The differences arise from different styles of spoken and written conversations, extra noise from grammatical errors, disfluencies or barge-ins, and speech recognition errors. To solve these issues, benchmark sets using audio were proposed.

\citet{hemphill1990atis} and \citet{bennett2002carnegie} proposed the speech dialogue benchmark which consists of fairly limited corpora focusing on travel reservations. Furthermore, there is no dialogue state annotation. 
Recently, with spoken dialog systems becoming popular, various speech dialogues benchmark sets have emerged that overcame these limitations, such as DSTC2 \citep{henderson2014second} and DSTC10 knowledge-grounded task-oriented dialogue modeling on spoken conversations track \citep{kim2021robust}. However, these benchmark sets were still insufficient to solve the speech-aware dialogue system because they had no dialog annotations, poor ASR system results, and/or a lack of audio data.

\section{Speech-aware Dialog Systems Task}

Here we provide background on the Speech-aware dialogue system technology track in DSTC11, which was introduced to overcome the challenges in implementing a good speech dialogue system with the previous speech dialog system benchmarks.

\subsection{Dataset} \label{dataset}

The dataset combines the Multi-domain Wizard-of-Oz (MultiWOZ) dataset \citep{kim2019efficient} and audio-related features.
The MultiWOZ dataset is a large-scale human-human conversational corpus spanning over seven domains, including about 10,000 multi-turn dialogues.
Each dialogue is rich in annotations such as ‘goal’, ‘meta-data’, ‘dialog act’, and user and system utterances.
Audio-related features are expressed in the following four kinds.
\begin{enumerate}
    \item Raw audio in the standard PCM format, 2 bytes per sample, at 16KHz sampling rate. Given as two types: TTS and human.
    \item Audio encoder output from the ASR system, consisting of 512-dimension vectors at a rate of 75 vectors per second.
    \item Transcripts from the ASR system (ASR hypothesis).
    \item Time alignment describing how the recognized words map to the encoder output sequences. For example, \textit{w:while t:2 w:in t:5 w:cam t:8 w:bridge t:11 w:â– t:15 w:i t:15 ...}
\end{enumerate}
The track tried to reduce the performance difference according to the ASR model by providing high-quality ASR results and feature vectors.

In addition, the state value of the dataset has been updated because the state value of the original MultiWOZ dataset was overlapped in the dev and test set compared to the train set, so fair evaluation could not be performed. Also, by changing the slot value in the conversation, the dataset was augmented 100 times.

Lastly, we used datasets with three kinds of audio as input to evaluate the spoken dialog system. The three types of datasets are classified according to the type of audio: \textit{TTS-Verbatim} generated by TTS, \textit{Human-Verbatim} by human speech, and \textit{Human-Paraphrased} by paraphrased human speech.
\begin{figure*}
    \centering
    \includegraphics[width=0.8\linewidth]{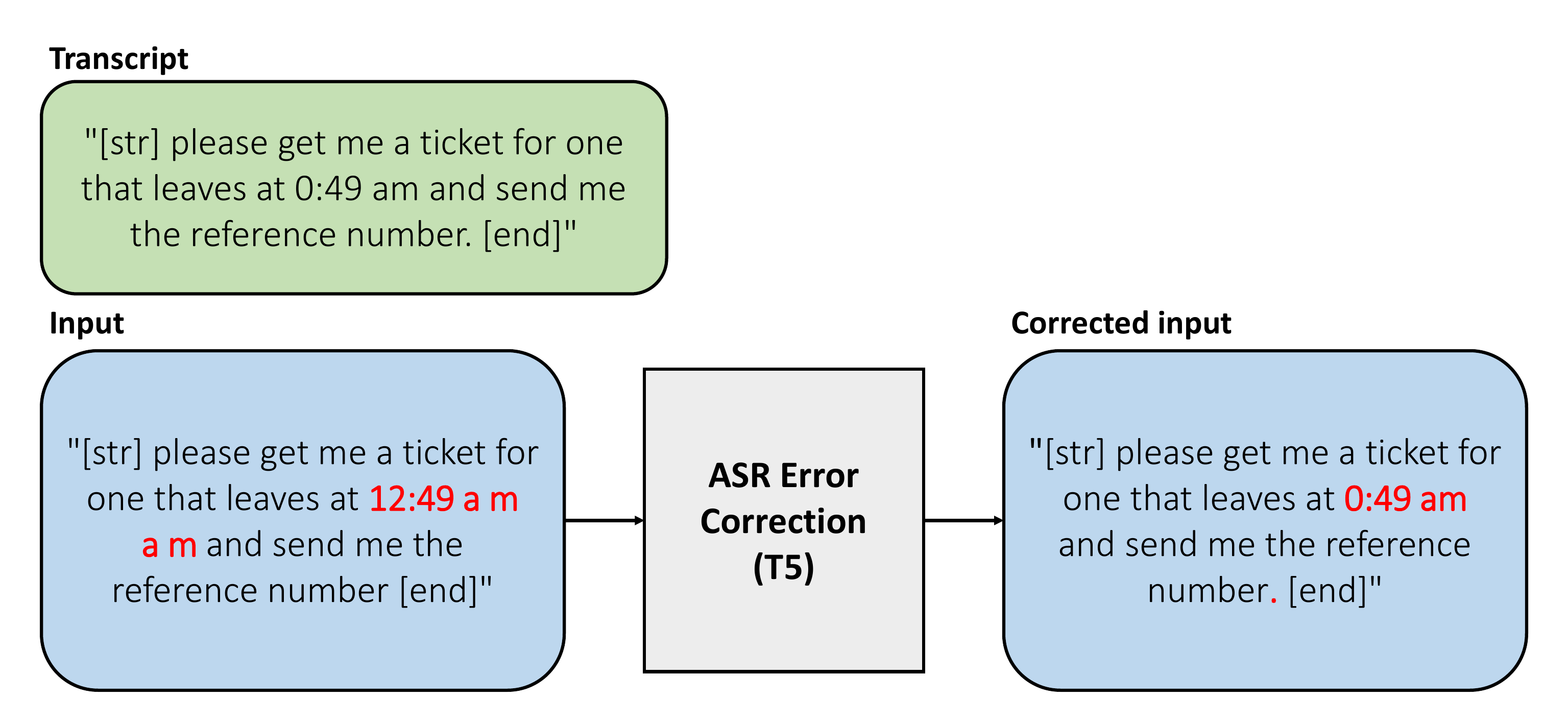}
    \caption{ ASR correction example (MultiWOZ \textit{mul0207.json}) }
    \label{fig:asr_text}
\end{figure*}
\subsection{Evaluation Methods}
The evaluation metric is dialogue state tracking\footnote{\url{https://github.com/jasonwu0731/trade-dst}}, one of the base evaluation metrics of the MultiWOZ dataset. Dialogue state tracking is a primary evaluation method that intuitively checks how well the dialogue model works for audio input. Joint goal accuracy(JGA) was used as the first evaluation metric, and slot error rate(SER) and F1 score were used as the second evaluation metric.
\section{Methods}
In this section, we describe our model that participated in the speech-aware dialogue systems technology challenge track in DSTC11. Figure~\ref{fig:model_structure} presents our model architecture. Our system consists of three modules: (1) ASR error correction, (2) Text Based Dialogue System, and (3) Post-processing. The overview of the process followed by our model is as follows:
\begin{enumerate}
    \item Take the given ASR result from the audio data set to create an input.
    \item Insert the input data into the \textbf{ASR error correction module} to obtain the text that corrected the ASR error.
    \item Output the value of the dialog state with the \textbf{dialog system} learned based on the text
    \item Run the \textbf{post processing module} to recover the dialog state value
\end{enumerate}

\subsection{ASR error correction}
In general, we use the ASR model to transcribe audio to text. However, errors arising from the ASR module can be propagated to the dialog system that uses them as input \citep{dutta2022error}, thus it is necessary to correct these ASR errors.
Since we have the ground truth text and corresponding ASR hypothesis data, we need to calibrate the ASR hypothesis to ground truth text.
As in previous studies, ASR errors can be improved by applying the n-best technique \citep{zhu2021improving}, but applying the n-best technique requires audio data for training, which can be very expensive. In addition, in the case of the ASR model, text converted through ASR can be inappropriate for certain downstream tasks. Therefore, instead of the n-best technique, we utilize an ASR error correction module to convert the input to a similar text in our ground truth transcript through a seq2seq-based model.

Figure~\ref{fig:asr_text} shows the flow of our ASR error correction module. By comparing the ground truth text and ASR hypothesis, it can be seen that there are many differences. There are special character disappearances, errors in certain formats such as time, as well as spelling errors. To fix this, we implemented a model that was trained to generate ground truth text from the ASR hypothesis based on T5 \citep{raffel2020exploring}. Therefore, our model can import recovered input instead of the ASR hypothesis which has a higher error compared to the original text.

\begin{figure*}
    \centering
    \includegraphics[width=0.8\linewidth]{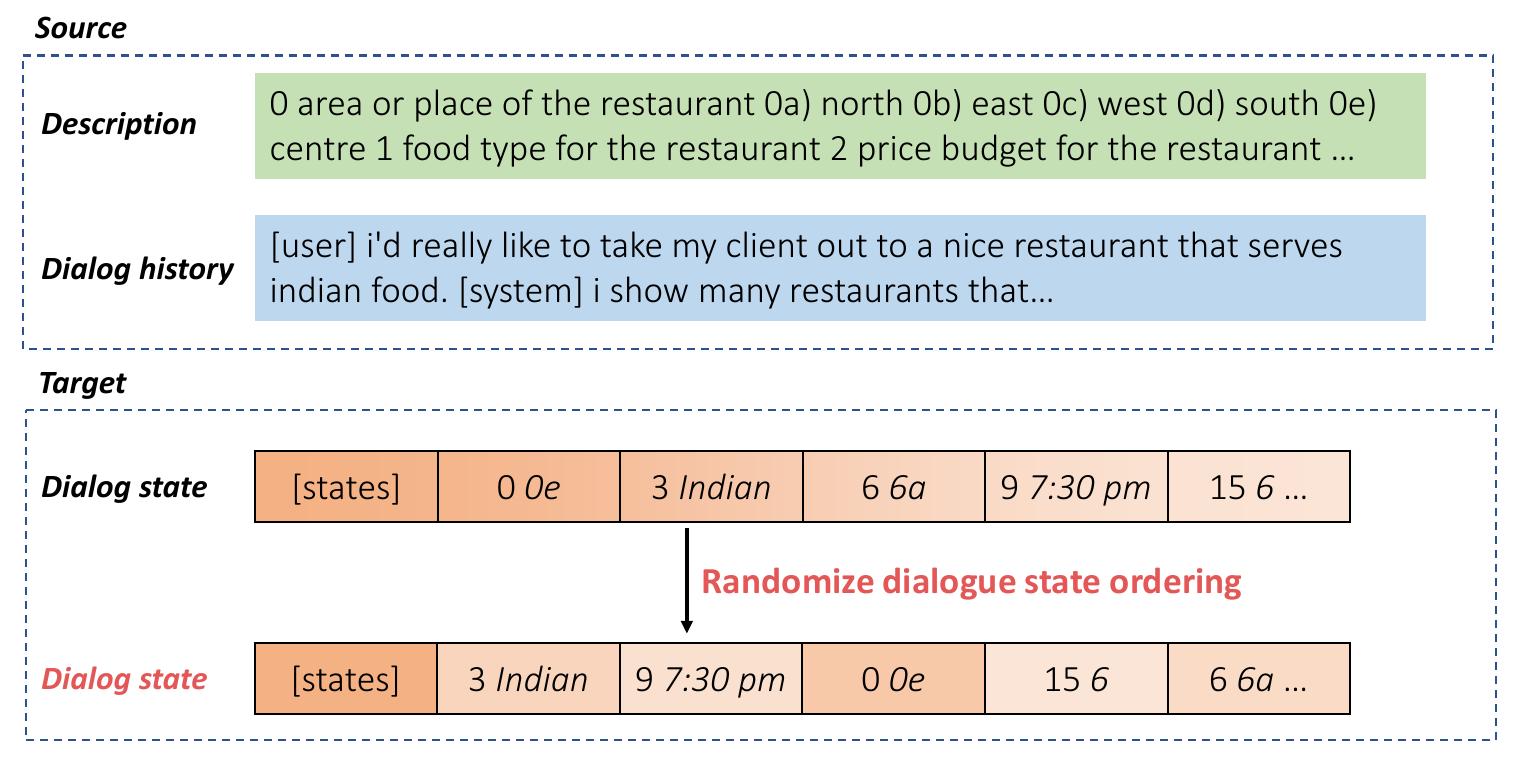}
    \caption{An example of DST model input. We applied a random ordering mechanism to D3ST-based input.}
    \label{fig:random_ordering}
\end{figure*}

\subsection{Inclusion of task description}
Previously, for TOD systems, there have been two main approaches: a modular approach to solve each task by module and the end-to-end approach \citep{madotto2018mem2seq}. The end-to-end approach could not show good performance due to the propagation of ASR errors in the spoken dialog system. In addition, the dataset provided on the track was difficult to perform well compared to the conventional MultiWOZ dataset. 

Therefore, we used Description-Driven DST \citep[D3ST;][]{zhao2022description} as a baseline to increase data efficiency by utilizing the description provided by MultiWOZ\footnote{\url{https://github.com/budzianowski/multiwoz}}. Also, we chose a T5-based generation model that showed higher efficiency in the DST Task.

Figure~\ref{fig:random_ordering} shows an example of the input structure of our DST model. First, we transformed the input data into D3ST format and applied an index-picking mechanism, handling categorical slots. Additionally, we applied random ordering techniques to target dialogue states for each example during training to prevent dependence on the order of predicted slots. When predicting a slot using a generation-based model, this method shows that the model does not depend on the order of slots, and as a result, the slot prediction performance increases.

\subsection{Post Processing}
There were many proper nouns such as hotel and restaurant names, but in the case of our model, which is using the ASR result as input, it was challenging for ASR error correction to correctly fix these proper nouns. In the previous study, the slot values were recovered by applying an encoder-decoder structure in a hierarchical manner \citep{ren2019scalable} or modified to a similar value using the Levenshtein distance ratio \citep{levenshtein1966binary}. To correct these proper nouns, we created a proper noun database based on Wikipedia data, and we created a module that uses word similarity to recover spelling error words. \\
There are various algorithms for calculating word similarity. We need to change the misrecognized word caused by ASR error to the correct word, but the spaCy library\footnote{\url{https://spacy.io/api/token\#similarity}} could not be used because it calculates the similarity of cosine in the vector of the word. In addition, the Gestalt pattern matching \citep{ratcliff1988pattern} algorithm is not symmetric, so it shows different similarities depending on the input order, and the performance is slow. Therefore, we applied the Levenshtein distance ratio\citep{levenshtein1966binary} algorithm that calculates similarity by editing (add/delete/update).
\section{Experiments}
We trained two T5\citep{raffel2020exploring} models for ASR error correction and the description dialogue system using huggingface. \texttt{T5-base}(220M) was used for ASR error correction and \texttt{T5-base, large}(770M) for the description dialog system. We used pre-trained weights by \citet{wolf2020transformers} and tokenized each sentence into sub-words using \texttt{T5Tokenizer}\footnote{\url{https://github.com/huggingface/transformers}}.

For fine-tuning, we used the AdamW optimizer \citep{loshchilov2017decoupled} with $\beta_1 = 0.9$, $\beta_2 = 0.999$, weight decay = 0.01 and a learning rate of 0.0001. We used a batch size of 8, an input length of 256, the output length of 256 for ASR error correction, and batch size of 6 (3 for \texttt{T5-large}), an input length of 1024, and an output length of 512 for the description dialogue system.

\subsection{ASR error correction Result}
\begin{table}[h!]
\small
\renewcommand{\arraystretch}{1.2}
\centering
\resizebox{0.9\columnwidth}{!}
    {
	\begin{tabular}{l|c|c}
		\thickhline
		Post Processing & ASR & ASR correction \\
		\hline \hline
		\textsc{None} & { 92.4\%} & {\textbf{57.5\%}} \\
		\hline
		\textsc{Removing special character} & {58.7\%} & {\textbf{22.0\%}}\\
		\thickhline
	\end{tabular}
    }
    \caption{Comparison of sentence error rate between the text of ASR without post-processing and of our ASR correction method. The sentence error rate is the number of incorrect sentences divided by the total number of sentences.}
\label{table:asr_correction}
\end{table}
We compared the text enhanced by our ASR correction method to the text generated by ASR without correction.
The result is summarized in Table~\ref{table:asr_correction}.
The first row shows the result without any post-processing text while the second row shows the result of removing special characters in the generated text.
The sentence error rate is the number of incorrect sentences divided by the total number of sentences.
We observe that our ASR correction method yields dramatic performance improvement, which is about a 35\% increase in sentence error rate in both cases.
This is because applying our ASR correction method makes the sentence from ASR more natural.
Qualitative results are presented in Figure~\ref{fig:asr_text}, which supports our claim.

\subsection{Main Results}

\begin{table}[h!]
    \begin{adjustbox}{width=\columnwidth}
    \begin{tabular}{ l|l|c|c  }
        \thickhline
        &Model + Feature&JGA&SER\\
        \hline
        \hline
        \multirow{4}{*}{Baseline}&D3ST (T5-base)&20.0&-\\
        &D3ST (T5-large)&21.1&-\\
        &D3ST (T5-large) + \textbf{D}&28.4&-\\
        &D3ST (T5-XXL) + \textbf{D}&33.7&-\\
        \hline
        \hline
        \multirow{6}{*}{Ours}&UBAR (gpt-2)&19.2&-\\
        &D3ST (T5-base)&23.3&31.7\\
        &D3ST (T5-base) + \textbf{A}&28.7&25.9\\
        &D3ST (T5-base) + \textbf{R}&26.2&-\\
        &D3ST (T5-base) + \textbf{R} + \textbf{P}&34.1&23.4\\
        &D3ST (T5-base) + \textbf{R} + \textbf{P} + \textbf{D}&40.3&19.0\\
        &D3ST (T5-base) + \textbf{R} + \textbf{P} + \textbf{D} + \textbf{A}&\textbf{41.6}&\textbf{18.3}\\
        \thickhline
    \end{tabular}
    \end{adjustbox}
    \caption{ The experiment results were trained on the original transcript and evaluated on the validation set. \textbf{A}: ASR Error Correction, \textbf{R}: Random Ordering, \textbf{P}: Post Processing, \textbf{D}: Data Augmentation}
    \label{tab:experiment_result}
\end{table}

Table~\ref{tab:experiment_result} shows the results of the experiment. This result is generated by training with \textit{original transcript} and evaluating it with \textit{tts-verbatim} of the validation set. The track organizers presented a baseline based on the \texttt{D3ST} model and showed differences in performance depending on the size of the backbone model and data augmentation.

Considering that the UBAR\citep{yang2021ubar} model operates in an end-to-end fashion, we implemented our backbone model based on D3ST\citep{zhao2022description} which can effectively solve DST Tasks.
On top of our backbone model, we carefully added each feature one by one to gradually increase the performance and identify its impact.

\textit{Random ordering} module allows us to avoid dependence on the order of slots predicted in the inference process, and \textit{Post Processing} module corrects prediction errors in words such as proper nouns. \textit{ASR Error Correction} module corrected ASR errors to text similar to the original transcript. Finally, using the \textit{Data Augmentation}, we were able to create high-performance models.

\begin{table}[h!]
    \begin{adjustbox}{width=\columnwidth}
    \begin{tabular}{ l|l|c|c  }
        \thickhline
        Model + Feature&Pretrain. Model (\# Params.)&JGA&SER\\
        \hline
        \hline
        \multirow{2}{*}{\shortstack[l]{D3ST\\\ \ + \textbf{R} + \textbf{P} + \textbf{D} + \textbf{A}}}&T5-base (220M)&41.6&\textbf{18.3}\\
            &T5-large (770M)&\textbf{42.4}&19.0\\
        \thickhline
    \end{tabular}
    \end{adjustbox}
    \caption{ Comparison of performance according to model size. \textbf{A}: ASR Error Correction, \textbf{R}: Random Ordering, \textbf{P}: Post Processing, \textbf{D}: Data Augmentation}
    \label{tab:comparison_of_model_size}
\end{table}
In addition, we conducted a performance comparison experiment according to the size of the T5 model. Table~\ref{tab:comparison_of_model_size} shows the result of the experiment. The performance of our model increased as the size of the base model increased, and \citet{zhao2022description} also showed that the performance difference between the \texttt{XXL} model and the \texttt{large} model was very large for the \texttt{D3ST} model. It would have been nice if we had experimented with a huge model such as \texttt{XXL}, but due to experimental constraints, we could only experiment with the \texttt{large} model. Finally, we submitted our final model based on the \texttt{large} model.

\begin{table*}[h!]
    \centering
    \begin{adjustbox}{width=0.95\textwidth}
    \begin{tabular}{ c|c|c|c|c|c|c  }
        \thickhline
        \multirow{2}{*}{System} & \multicolumn{3}{c|}{Joint Goal Accuracy} & \multicolumn{3}{c}{Slot Error Rate}\\
        &TTS-verbatim&Human-verbatim&Human-paraphrased&TTS-verbatim&Human-verbatim&Human-paraphrased\\
        \hline
        \hline
        F-p&44.0&39.5&37.9&17.1&20.0&20.4\\
        F-s&40.4&36.1&34.3&19.2&21.9&22.4\\
        \textbf{C-p (ours)}&\textbf{40.2}&\textbf{31.9}&\textbf{31.8}&\textbf{20.9}&\textbf{28.1}&\textbf{27.2}\\
        A-s&37.7&30.1&30.7&20.3&26.9&26.2\\
        C-s&33.1&28.6&28.1&25.0&28.7&29.5\\
        D-s&30.3&23.5&23.2&26.6&36.5&35.1\\
        B-p&27.3&23.9&22.6&26.2&30.0&30.6\\
        D-p&30.3&23.5&23.2&28.0&36.7&36.0\\
        A-p&21.9&21.2&20.0&32.8&33.5&33.8\\
        B-s&22.4&19.2&18.3&28.7&32.2&32.6\\
        E-p&21.3&20.0&18.2&35.1&35.5&35.3\\
        \thickhline
    \end{tabular}
    \end{adjustbox}
    \caption{ Official results for test submissions by \texttt{DSTC11-Track3}. The text in bold indicates our model. The first letter of System means \textbf{Team}, and the second letter means System, \textbf{p} is \textbf{Primary System} and \textbf{s} is \textbf{Secondary System}.}
    \label{tab:official_result}
\end{table*}

Table~\ref{tab:official_result} is the official results of the test submission by the participants. A total of 6 teams submitted, and each team could submit up to 2 systems, so a total of 11 systems were submitted. We submitted a model that recorded 42.4 in the validation set. Finally, our model achieved third place, with JGA 40.2 for \textit{tts-verbatim} in the challenge.

In particular, our model showed a small gap between \textit{human-verbatim} and \textit{human-paraphrased} data because it was implemented through the \textit{ASR Error Correction} module in a flexible form for sentence structure.

\subsection{Worst-Case Analysis}
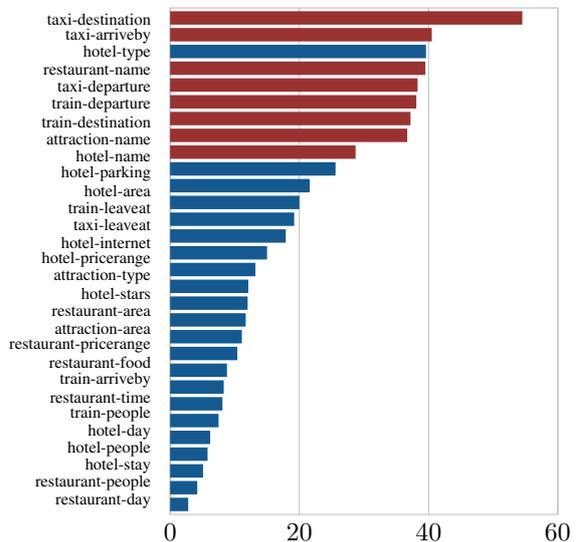
\begin{figure}[h]
\resizebox{\columnwidth}{!}{%
\begin{tikzpicture}
    \begin{axis}[%
    width=0.95\columnwidth,
    xbar, y=-0.25cm, bar width=0.2cm, enlarge y limits={abs=0.15cm}, xmin=0, xmax=60,
    symbolic y coords={
        test, hospitaldepartment, taxidestination, taxiarriveby, hoteltype, restaurantname, taxideparture, traindeparture,
        traindestination, attractionname, hotelname, hotelparking, hotelarea, trainleaveat, taxileaveat, hotelinternet,
        hotelpricerange, attractiontype, hotelstars, restaurantarea, attractionarea, restaurantpricerange,
        restaurantfood, trainarriveby, restauranttime, trainpeople, hotelday, hotelpeople, hotelstay,
        restaurantpeople, restaurantday, trainday
    },
    y tick label style={align=right,font=\scriptsize},
    ytick distance=1.025, 
    yticklabels={
        test, hospital-department, taxi-destination, taxi-arriveby, hotel-type, restaurant-name, taxi-departure, train-departure,
        train-destination, attraction-name, hotel-name, hotel-parking, hotel-area, train-leaveat, taxi-leaveat,
        hotel-internet, hotel-pricerange, attraction-type, hotel-stars, restaurant-area, attraction-area,
        restaurant-pricerange, restaurant-food, train-arriveby, restaurant-time, train-people,
        hotel-day, hotel-people, hotel-stay, restaurant-people, restaurant-day, train-day
    },
    xmajorgrids,
    axis line style={lightgray},
    major tick style={draw=none},
    nodes near coords,
    point meta=explicit symbolic,
    /pgf/bar shift={0pt},   
    ]
    \addplot [fill={rgb:red,170;green,55;blue,55},draw=none] coordinates {
        (142.9,hospitaldepartment) (54.5,taxidestination) (40.5,taxiarriveby) (39.5,restaurantname) (38.3,taxideparture) (38.1,traindeparture) (37.2,traindestination) (36.7,attractionname) (28.7,hotelname)
    };
    \addplot [fill={rgb:red,23;green,105;blue,170},draw=none] coordinates {
        (39.6,hoteltype) (25.6,hotelparking) (21.6,hotelarea) (20.0,trainleaveat) (19.2,taxileaveat) (17.9,hotelinternet) (15.0,hotelpricerange)
        (13.2,attractiontype) (12.1,hotelstars) (12.0,restaurantarea) (11.7,attractionarea) (11.1,restaurantpricerange) (10.4,restaurantfood) (8.8,trainarriveby) (8.3,restauranttime) (8.1,trainpeople) (7.5,hotelday)
        (6.2,hotelpeople) (5.8,hotelstay) (5.1,restaurantpeople) (4.2,restaurantday) (2.8,trainday)
    };
    \end{axis}
\end{tikzpicture}
}
\caption{Slots error rate per each slot. Most slots with high slot error rates are slots with proper nouns as slot values, for example, \textit{taxi-destination} (54.5\%), \textit{taxi-arriveby} (40.5\%), and \textit{restaurant-name} (39.5\%). Red-colored slots contain proper nouns.}
\label{fig:slot_error_rate_per_slot_name}
\end{figure}
We hypothesize several approaches that potentially improve the performance of our model.
Figure~\ref{fig:slot_error_rate_per_slot_name} shows the distribution of slot error rates for each slot. Slots containing proper nouns, such as \textit{hotel-name} and \textit{restaurant-name}, showed high slot error rates. Like \texttt{DSTC11-Track3}, in the case of a Dialogue system that uses ASR results as input, there can be many problems with proper nouns. We tried to correct these errors through post-processing, but there was still a lot of ambiguity because we have no dataset or ontology for proper nouns in the model. So we think that defining the appropriate ontology of the system and post-processing will lead to even better performance.

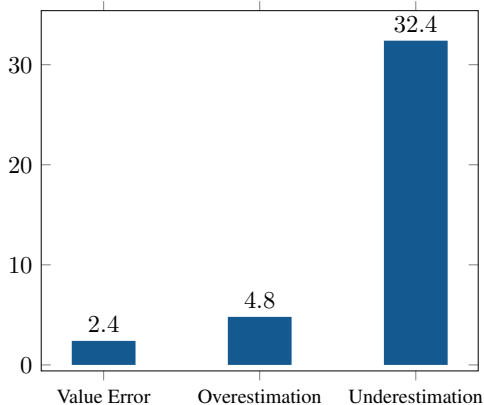
\begin{figure}
\centering
\resizebox{0.85\columnwidth}{!}{%
    \begin{tikzpicture}
        \begin{axis}[
            ybar, bar width=1cm, 
            symbolic x coords={v, o, u},
            legend pos = north west, 
            nodes near coords, 
            enlarge x limits=0.2, 
            every node near coord/.append style={font=\normalsize},
            xtick=data, x tick label style={align=right,font=\small},
            xticklabels={
                Value Error, Overestimation, Underestimation
            },
        ] 
        \addplot [fill={rgb:red,23;green,105;blue,170},draw=none] coordinates {(v, 2.4) (o, 4.8) (u, 32.4)};
        \end{axis} 
    \end{tikzpicture} 
}
\caption{The cause of the error in \textit{hotel-type} slot. Most of the reasons for the error in \textit{hotel-type} slot were believed to have been caused by underestimation.}
\label{fig:error_rate_by_cause}
\end{figure}

Figure~\ref{fig:error_rate_by_cause} shows the distribution of the error cause of the \textit{hotel-type} slot, which showed a high slot error rate even though the slot does not include proper nouns. As a result of the analysis, most of the errors appeared to be caused by underestimation. We believe that this is caused by many cases in which hotel types are included in the hotel name, such as \textit{guest house} and \textit{hotel}. Therefore, we think that it would be good to solve the inference about the hotel type by separating the hotel name and the hotel type.

\section{Conclusion}
In this paper, we proposed an ASR dialogue pipeline model to solve the speech-aware dialogue system technology track in DSTC11. First, we reduce the gap when applying spoken data to text-based dialogue systems by modifying ASR results that differ from the original text such as time formats or special characters through the ASR error correction module. Secondly, the text-based dialogue system, a model including descriptions with several techniques, was able to get good performance even with low input data feature quality of audio data. Lastly, by recovering errors such as proper nouns that cannot be caught in ASR error correction through post-processing, a dialogue system that produces high performance even on the speech database was created. In this way, we have created a model that can overcome the limitations of speech in several parts, and in the case of T5, although the model is very sensitive to parameter size, we can confirm that it is a model with high potential for development by performing well even in a small size model.

\section*{Acknowledgments and Disclosure of Funding}
This work was supported by Electronics and Telecommunications Research Institute(ETRI) grant funded by the Korean government (22ZS1100, Core Technology Research for Self-Improving Integrated Artificial Intelligence System),  Institute of Information \& communications Technology Planning \& Evaluation(IITP) grant funded by the Korea government(MSIT) (No.2020-0-00940, Foundations of Safe Reinforcement Learning and Its Applications to Natural Language Processing; No.2019-0-00075, Artificial Intelligence Graduate School Program (KAIST)).
\\~~
\bibliography{acl_latex}

\begin{thebibliography}{22}
\expandafter\ifx\csname natexlab\endcsname\relax\def\natexlab#1{#1}\fi

\bibitem[{Bennett and Rudnicky(2002)}]{bennett2002carnegie}
Christina Bennett and Alexander Rudnicky. 2002.
\newblock The carnegie mellon communicator corpus.

\bibitem[{Brown et~al.(2020)Brown, Mann, Ryder, Subbiah, Kaplan, Dhariwal,
  Neelakantan, Shyam, Sastry, Askell et~al.}]{brown2020language}
Tom Brown, Benjamin Mann, Nick Ryder, Melanie Subbiah, Jared~D Kaplan, Prafulla
  Dhariwal, Arvind Neelakantan, Pranav Shyam, Girish Sastry, Amanda Askell,
  et~al. 2020.
\newblock Language models are few-shot learners.
\newblock \emph{Advances in neural information processing systems},
  33:1877--1901.

\bibitem[{Dutta et~al.(2022)Dutta, Jain, Maheshwari, Ramakrishnan, and
  Jyothi}]{dutta2022error}
Samrat Dutta, Shreyansh Jain, Ayush Maheshwari, Ganesh Ramakrishnan, and
  Preethi Jyothi. 2022.
\newblock Error correction in asr using sequence-to-sequence models.
\newblock \emph{arXiv preprint arXiv:2202.01157}.

\bibitem[{Hemphill et~al.(1990)Hemphill, Godfrey, and
  Doddington}]{hemphill1990atis}
Charles~T Hemphill, John~J Godfrey, and George~R Doddington. 1990.
\newblock The atis spoken language systems pilot corpus.
\newblock In \emph{Speech and Natural Language: Proceedings of a Workshop Held
  at Hidden Valley, Pennsylvania, June 24-27, 1990}.

\bibitem[{Henderson et~al.(2014)Henderson, Thomson, and
  Williams}]{henderson2014second}
Matthew Henderson, Blaise Thomson, and Jason~D Williams. 2014.
\newblock The second dialog state tracking challenge.
\newblock In \emph{Proceedings of the 15th annual meeting of the special
  interest group on discourse and dialogue (SIGDIAL)}, pages 263--272.

\bibitem[{Khan et~al.(2015)Khan, Robichaud, Crook, and
  Sarikaya}]{khan2015hypotheses}
Omar~Zia Khan, Jean-Philippe Robichaud, Paul~A Crook, and Ruhi Sarikaya. 2015.
\newblock Hypotheses ranking and state tracking for a multi-domain dialog
  system using multiple asr alternates.
\newblock In \emph{Sixteenth Annual Conference of the International Speech
  Communication Association}.

\bibitem[{Kim et~al.(2021)Kim, Liu, Jin, Papangelis, Gopalakrishnan,
  Hedayatnia, and Hakkani-T{\"u}r}]{kim2021robust}
Seokhwan Kim, Yang Liu, Di~Jin, Alexandros Papangelis, Karthik Gopalakrishnan,
  Behnam Hedayatnia, and Dilek Hakkani-T{\"u}r. 2021.
\newblock “how robust ru?”: Evaluating task-oriented dialogue systems on
  spoken conversations.
\newblock In \emph{2021 IEEE Automatic Speech Recognition and Understanding
  Workshop (ASRU)}, pages 1147--1154. IEEE.

\bibitem[{Kim et~al.(2019)Kim, Yang, Kim, and Lee}]{kim2019efficient}
Sungdong Kim, Sohee Yang, Gyuwan Kim, and Sang-Woo Lee. 2019.
\newblock Efficient dialogue state tracking by selectively overwriting memory.
\newblock \emph{arXiv preprint arXiv:1911.03906}.

\bibitem[{Levenshtein et~al.(1966)}]{levenshtein1966binary}
Vladimir~I Levenshtein et~al. 1966.
\newblock Binary codes capable of correcting deletions, insertions, and
  reversals.
\newblock In \emph{Soviet physics doklady}, volume~10, pages 707--710. Soviet
  Union.

\bibitem[{Loshchilov and Hutter(2017)}]{loshchilov2017decoupled}
Ilya Loshchilov and Frank Hutter. 2017.
\newblock Decoupled weight decay regularization.
\newblock \emph{arXiv preprint arXiv:1711.05101}.

\bibitem[{Madotto et~al.(2020)Madotto, Liu, Lin, and
  Fung}]{madotto2020language}
Andrea Madotto, Zihan Liu, Zhaojiang Lin, and Pascale Fung. 2020.
\newblock Language models as few-shot learner for task-oriented dialogue
  systems.
\newblock \emph{arXiv preprint arXiv:2008.06239}.

\bibitem[{Madotto et~al.(2018)Madotto, Wu, and Fung}]{madotto2018mem2seq}
Andrea Madotto, Chien-Sheng Wu, and Pascale Fung. 2018.
\newblock Mem2seq: Effectively incorporating knowledge bases into end-to-end
  task-oriented dialog systems.
\newblock \emph{arXiv preprint arXiv:1804.08217}.

\bibitem[{Raffel et~al.(2020)Raffel, Shazeer, Roberts, Lee, Narang, Matena,
  Zhou, Li, Liu et~al.}]{raffel2020exploring}
Colin Raffel, Noam Shazeer, Adam Roberts, Katherine Lee, Sharan Narang, Michael
  Matena, Yanqi Zhou, Wei Li, Peter~J Liu, et~al. 2020.
\newblock Exploring the limits of transfer learning with a unified text-to-text
  transformer.
\newblock \emph{J. Mach. Learn. Res.}, 21(140):1--67.

\bibitem[{Ratcliff and Metzener(1988)}]{ratcliff1988pattern}
John~W Ratcliff and David~E Metzener. 1988.
\newblock Pattern-matching-the gestalt approach.
\newblock \emph{Dr Dobbs Journal}, 13(7):46.

\bibitem[{Ren et~al.(2019)Ren, Ni, and McAuley}]{ren2019scalable}
Liliang Ren, Jianmo Ni, and Julian McAuley. 2019.
\newblock Scalable and accurate dialogue state tracking via hierarchical
  sequence generation.
\newblock \emph{arXiv preprint arXiv:1909.00754}.

\bibitem[{Shah et~al.(2019)Shah, Gupta, Fayazi, and
  Hakkani-Tur}]{shah2019robust}
Darsh~J Shah, Raghav Gupta, Amir~A Fayazi, and Dilek Hakkani-Tur. 2019.
\newblock Robust zero-shot cross-domain slot filling with example values.
\newblock \emph{arXiv preprint arXiv:1906.06870}.

\bibitem[{Thoppilan et~al.(2022)Thoppilan, De~Freitas, Hall, Shazeer,
  Kulshreshtha, Cheng, Jin, Bos, Baker, Du et~al.}]{thoppilan2022lamda}
Romal Thoppilan, Daniel De~Freitas, Jamie Hall, Noam Shazeer, Apoorv
  Kulshreshtha, Heng-Tze Cheng, Alicia Jin, Taylor Bos, Leslie Baker, Yu~Du,
  et~al. 2022.
\newblock Lamda: Language models for dialog applications.
\newblock \emph{arXiv preprint arXiv:2201.08239}.

\bibitem[{Tian et~al.(2021)Tian, Huang, He, Lin, Bao, He, Huang, Ju, Zhang, Xie
  et~al.}]{tian2021tod}
Xin Tian, Xinxian Huang, Dongfeng He, Yingzhan Lin, Siqi Bao, Huang He, Liankai
  Huang, Qiang Ju, Xiyuan Zhang, Jian Xie, et~al. 2021.
\newblock Tod-da: Towards boosting the robustness of task-oriented dialogue
  modeling on spoken conversations.
\newblock \emph{arXiv preprint arXiv:2112.12441}.

\bibitem[{Wolf et~al.(2020)Wolf, Debut, Sanh, Chaumond, Delangue, Moi, Cistac,
  Rault, Louf, Funtowicz et~al.}]{wolf2020transformers}
Thomas Wolf, Lysandre Debut, Victor Sanh, Julien Chaumond, Clement Delangue,
  Anthony Moi, Pierric Cistac, Tim Rault, R{\'e}mi Louf, Morgan Funtowicz,
  et~al. 2020.
\newblock Transformers: State-of-the-art natural language processing.
\newblock In \emph{Proceedings of the 2020 conference on empirical methods in
  natural language processing: system demonstrations}, pages 38--45.

\bibitem[{Yang et~al.(2021)Yang, Li, and Quan}]{yang2021ubar}
Yunyi Yang, Yunhao Li, and Xiaojun Quan. 2021.
\newblock Ubar: Towards fully end-to-end task-oriented dialog system with
  gpt-2.
\newblock In \emph{Proceedings of the AAAI Conference on Artificial
  Intelligence}, volume~35, pages 14230--14238.

\bibitem[{Zhao et~al.(2022)Zhao, Gupta, Cao, Yu, Wang, Lee, Rastogi, Shafran,
  and Wu}]{zhao2022description}
Jeffrey Zhao, Raghav Gupta, Yuan Cao, Dian Yu, Mingqiu Wang, Harrison Lee,
  Abhinav Rastogi, Izhak Shafran, and Yonghui Wu. 2022.
\newblock Description-driven task-oriented dialog modeling.
\newblock \emph{arXiv preprint arXiv:2201.08904}.

\bibitem[{Zhu et~al.(2021)Zhu, Liu, Liu, and Lin}]{zhu2021improving}
Linchen Zhu, Wenjie Liu, Linquan Liu, and Edward Lin. 2021.
\newblock Improving asr error correction using n-best hypotheses.
\newblock In \emph{2021 IEEE Automatic Speech Recognition and Understanding
  Workshop (ASRU)}, pages 83--89. IEEE.

\end{thebibliography}
\bibliographystyle{acl_natbib}




\end{document}